\documentclass[11pt]{article}

\usepackage[final]{acl}

\usepackage{times}
\usepackage{latexsym}

\usepackage[T1]{fontenc}
\usepackage[utf8]{inputenc}

\usepackage{microtype}

\usepackage{inconsolata}

\usepackage{graphicx}

\usepackage{hyperref}
\usepackage{url}
\usepackage{booktabs}
\usepackage{multirow}
\usepackage{graphicx} 
\usepackage{array}
\usepackage{tabularx}
\usepackage{subcaption}
\usepackage{wrapfig}
\usepackage[most]{tcolorbox}
\usepackage{listings}
\usepackage{xcolor}
\usepackage{makecell}
\usepackage[table]{xcolor}
\usepackage{algorithm}
\usepackage{amsmath}
\usepackage{tcolorbox}
\usepackage{algpseudocode}

\definecolor{softblue}{RGB}{200, 225, 240}   % 柔和蓝色（更淡）
\definecolor{warmpink}{RGB}{255, 205, 210}   % 温暖粉色（浅一些）
\definecolor{mintgreen}{RGB}{200, 235, 200}  % 薄荷绿（偏淡）
\definecolor{lightyellow}{RGB}{255, 255, 200}% 浅黄色（柔和）
\definecolor{lavender}{RGB}{240, 240, 250}   % 薰衣草紫（更浅紫灰感）
\definecolor{peach}{RGB}{255, 230, 210}      % 桃子橙（淡柔）

\definecolor{lightgray}{rgb}{0.95,0.95,0.95}
\tcbuselibrary{breakable}

\newtcolorbox{dialogbox}{
  breakable, 
  colback=gray!10,
  colframe=black,
  boxrule=0.5pt,
  arc=4pt,
  boxsep=0.25ex,
  left=0.5ex,
  right=0.5ex,
  top=0.75ex,
  bottom=0.75ex,
  fontupper=\normalsize,
}

\title{Empowering GUI Agents via Autonomous Experience Exploration and Hindsight Experience Utilization for Task Planning}

\author{
 \textbf{Tianyi Men\textsuperscript{1,2}},
 \textbf{Zhuoran Jin\textsuperscript{1,2}},
 \textbf{Pengfei Cao\textsuperscript{1,2}},
 \textbf{Yubo Chen\textsuperscript{1,2}},
 \textbf{Kang Liu\textsuperscript{1,2}},
 \textbf{Jun Zhao\textsuperscript{1,2,\dag}}
\\
 \textsuperscript{1}The Key Laboratory of Cognition and Decision Intelligence for Complex Systems,\\
 Institute of Automation, Chinese Academy of Sciences, Beijing, China\\
 \textsuperscript{2}School of Artificial Intelligence, University of Chinese Academy of Sciences, Beijing, China
\\
 \small{
   \{tianyi.men, zhuoran.jin, pengfei.cao,  yubo.chen, kliu, jzhao\}@nlpr.ia.ac.cn
 }
}

\begin{document}
\maketitle
\def\thefootnote{\dag}\footnotetext{Corresponding author.}\def\thefootnote{\arabic{footnote}}
\begin{abstract}

Multimodal web agents can assist humans in operating repetitive GUI tasks, where effective task planning is essential for decomposing complex tasks into executable actions. While small open‑source MLLMs are cost‑efficient and privacy-preserving compared with commercial large models, they suffer from weak planning and limited cross‑website generalization. To address these limitations, we introduce the planning experience exploration and utilization (PEEU) method, which autonomously explores environments to discover experiences and utilizes hindsight experience to synthesize strictly aligned, high-level training data. To quantitatively analyze the generalization behaviors driving this performance, we propose the task decomposition hierarchical analysis framework (TDHAF) to systematically study compositional generalization across three task granularities: low, middle and high levels. Our analysis reveals that mastering low‑level atomic skills does not guarantee high‑level planning competence, while high‑level task training yields stronger OOD generalization. Experiments on real-world benchmarks demonstrate PEEU's superior effectiveness: our 7B model achieves 30.6\% accuracy, outperforming the much larger Qwen2.5-VL-32B model. These demonstrate constructing hindsight high‑level tasks and leveraging experiences is crucial for OOD planning abilities of small MLLMs.

\end{abstract}

\section{Introduction}

\begin{figure}[!t]
\begin{center}
\includegraphics[width=\linewidth]{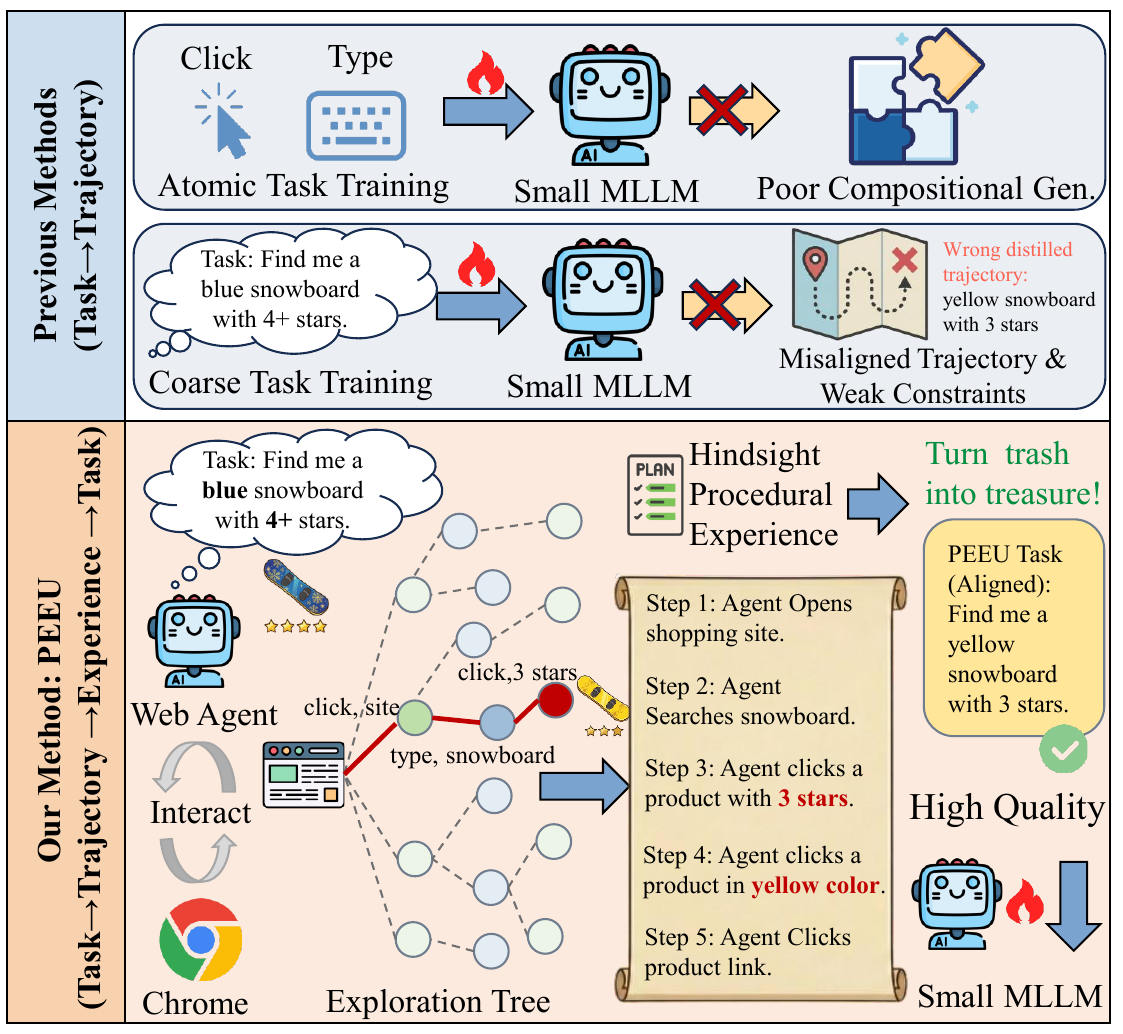}
\end{center}
\caption{The overview of planning experience exploration and utilization method.}
\label{fig:introduction}
\vspace{-20pt}
\end{figure}

The multimodal web agent is an attractive solution, which can assist humans in operating on unfamiliar websites and handling repetitive GUI tasks~\citep{wang2024gui,ning2025survey,tang2025survey}. The core ability of the agent is task planning, which enables it to decompose a complex task into executable actions~\citep{li2025perception,cao2025large,wei2025plangenllms}. Due to the high interaction costs and privacy risks of commercial large models, using small open-source multimodal large language models (MLLMs) is a promising approach~\citep{belcak2025small}. However, small MLLMs exhibit weak planning ability and limited generalization. Thus, enhancing their planning with limited data is urgent~\citep{he2024webvoyager}. In comparison, humans can make plans by utilizing experiences from interaction and exploration with the environment~\citep{ross1989some,anderson2013architecture}. Inspired by the human learning process, agents should (1) autonomously set their own learning goals in the environment and improve their abilities through interaction and exploration, and (2) summarize and utilize hindsight experiences from the past to guide future decisions~\citep{silver2025welcome,cai2025building}.
%zhang2025landscape

Recent studies focus on utilizing experiences in the post-training stage to train models. As shown in Figure~\ref{fig:introduction}, these approaches can be categorized into two main streams:
(1) Training with atomic-level tasks~\citep{gu2024your,fan2025gui}. These methods compare changes before and after environment observations to extract experiences. The experiences are then used to synthesize atomic-level tasks such as clicking, typing, and scrolling to train the model. However, it remains unclear whether training on atomic-level tasks can effectively generalize to high-level tasks. Hence, it is urgent to propose a framework to study the compositional generalization of web agent task planning.
(2) Training with coarse high-level tasks~\citep{logeswaranscaling,trabucco2025insta}. These methods leverage task-based exploration trajectories to train the model with coarse high-level tasks, like finding a snowboard with constraints. However, trajectories of coarse high-level tasks suffer from misalignment and a lack of stricter constraints. This limits the generalization ability in high-level tasks. Therefore, it is necessary to develop a method to synthesize trajectories that are better aligned and strictly constrained by environments.

To address these limitations while ensuring a fair comparison using the same scale data, we propose the \textbf{planning experience exploration and utilization} method (\textbf{PEEU}), as shown in Figure~\ref{fig:introduction}. Distinct from previous methods that rely on brute-force search to match trajectories with pre-defined goals, we leverage hindsight to inversely align tasks to the collected trajectories, thereby significantly enhancing the quality of high-level data. The framework consists of two stages: planning tree exploration and planning experience utilization. (1) In the \textbf{planning tree exploration} stage, the exploration model autonomously sets goals adapted to the functional characteristics of diverse websites, and then conducts goal-driven exploration in the unfamiliar environment to construct an exploration tree. (2) In the \textbf{planning experience utilization} stage, trajectories are summarized to extract valuable experiences. These experiences are then used to create better aligned and constrained pairs of tasks and trajectories. We evaluate PEEU on seven unseen real-world websites. Under a strictly controlled setting with identical data scales for all methods, PEEU demonstrates superior cross-website generalization. PEEU based on Qwen2.5-VL-7B reaches 30.6\% accuracy, marking a significant improvement over the Instruct Model's performance of 7.8\%.

To further validate the advantage of high-level tasks over atomic-level tasks, we propose the \textbf{task decomposition hierarchical analysis framework} (\textbf{TDHAF}). Our analysis confirms that mastering atomic skills is insufficient for complex planning, thereby validating PEEU's emphasis on high-level experience. This framework first defines three levels of task granularity: \textbf{low-level} tasks, \textbf{mid-level} tasks, and \textbf{high-level} tasks. It further distinguishes between two types of generalization: in-domain (\textbf{ID}) and out-of-domain (\textbf{OOD}). Building on this taxonomy, we analyze from three perspectives: (1) \textbf{ID bottom-up generalization}: whether low-level tasks can generalize to high-level tasks in-domain. (2) \textbf{ID top-down generalization}: whether high-level tasks can generalize to low-level tasks in-domain. (3) \textbf{OOD multi-level generalization}: what granularity of tasks is better for out-of-domain generalization.
The experiments demonstrate following conclusions: (1) Mastering individual low-level tasks does not necessarily imply mastery of the corresponding high-level task. (2) Using high-level tasks makes it easier to generalize downwards in-domain with greater overall coverage. (3) Using high-level task training can enable the model to acquire stronger generalization capabilities for multi-level tasks in OOD. 
Overall, experiments show that in post-training stage, using low-level tasks cannot effectively generalize to high-level tasks.

In summary, our contributions are as follows: (1) We propose the \textbf{planning experience exploration and utilization} method (\textbf{PEEU}), which can autonomously explores and effectively utilizes experiences to enhance the planning generalization abilities of web agents. (2) We propose the \textbf{task decomposition hierarchical analysis framework} (\textbf{TDHAF}) to analyze the compositional generalization ability of models in multimodal web navigation task planning scenarios. (3) PEEU improves cross-website OOD generalization in real online multimodal web navigation tasks, outperforming previous methods across different model scales with the same data scale and training settings.

\section{Planning Experience Exploration and Utilization Method}
In this section, we introduce the planning experience exploration and utilization method. This is an automatic exploration learning framework that first sets goals adaptively and explores in unfamiliar websites. Then it extracts planning experiences from trajectories and uses them to build aligned and constrained training data. Users only need to provide a URL to be explored, and the framework can freely explore the website, extract and summarize experiences, and then build better aligned and constrained data to train small MLLMs, achieving cross-website generalization capabilities.

\begin{figure*}[h]
\begin{center}
\includegraphics[width=\linewidth]{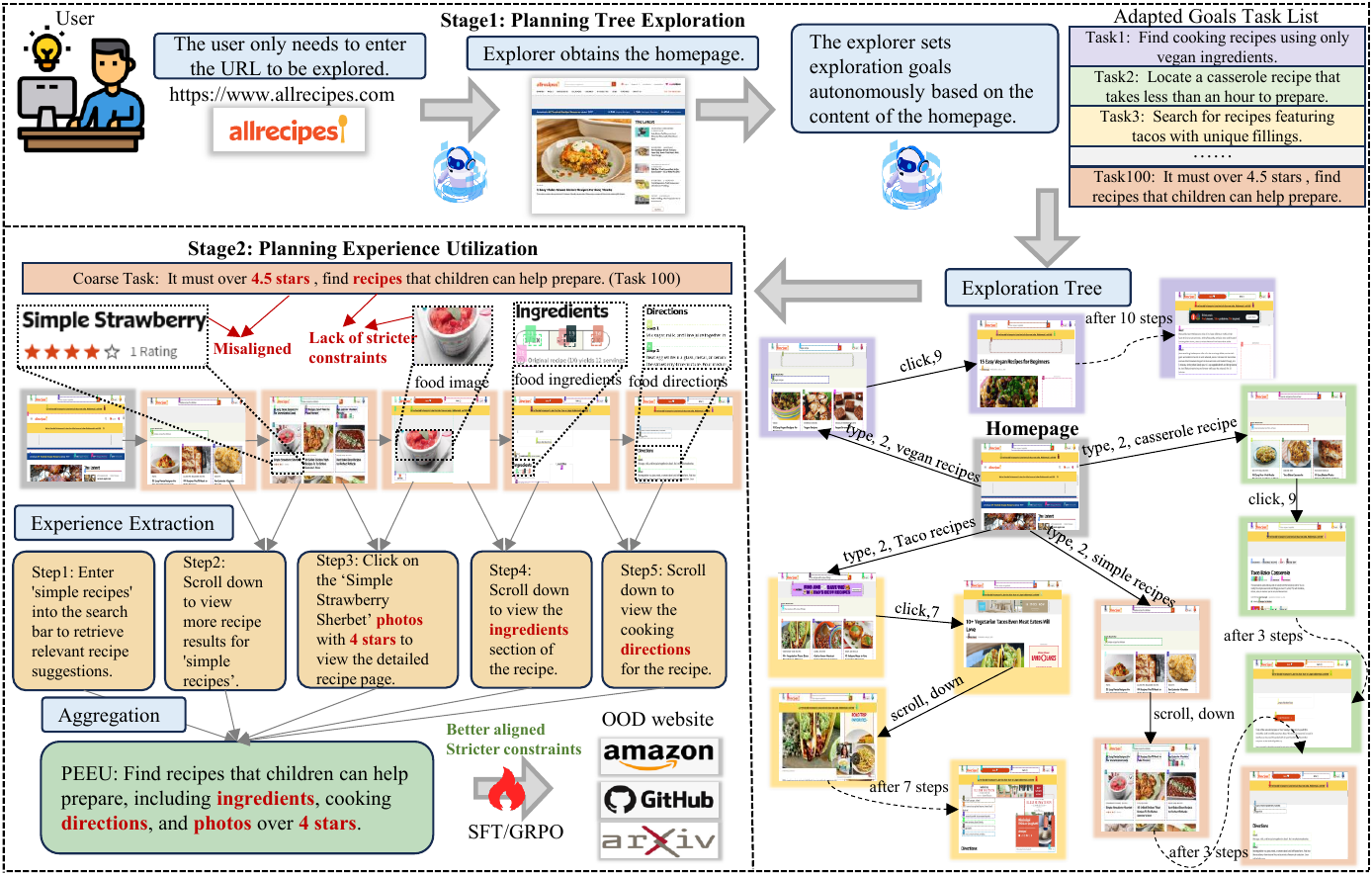}
\end{center}
\caption{An overview of planning experience exploration and utilization method with two stages.}
\label{fig:Method}
\vspace{-10pt}
\end{figure*}

\subsection{Method}

The framework is divided into two stages: \textbf{planning tree exploration} and \textbf{planning experience utilization}, as shown in Figure~\ref{fig:Method}.
All prompts are shown in Appendix~\ref{appendix:PEEU Prompt}.

\paragraph{Planning Tree Exploration.}
The autonomous agent requires a shift from passive learning to autonomous learning. It requires self-driven tasks and self-execution exploration. For the self-driven tasks stage, given a website URL, the exploration agent interacts with the homepage $s_0$ (obtained from the URL) through the MLLM $M$ to generate a basic task list $\mathcal{D} = \{ d_1, d_2, \dots, d_n \}$, where each task $d_i$ represents a task to be explored. This process can be expressed as:

\begin{equation}
\mathcal{D} = M(s_0, \text{URL}).
\end{equation}

Subsequently, for the self-execution exploration stage, the agent performs autonomous exploration based on the task list $\mathcal{D}$, the environment $\text{Env}$ (with basic URL as entry point), generating a directed exploration tree $\mathcal{R} = (V, E)$ rooted at the homepage, where $V$ is the set of website screens, $E$ is the set of actions between these observations. The exploration process is implemented as:
\begin{equation}
\mathcal{R} = \text{Explore}(M, \mathcal{D}, \text{Env}, \text{URL}).
\end{equation}

This tree can be expanded into interleaved trajectories of observations and actions, where all trajectories share the same root node. Formally, let $\tau=\{(s_0,a_0),\ldots,(s_m,a_m)\}$ denote a trajectory, where $s_0$ is the shared root state (homepage). $a_t \in \mathcal{A}$ represents the action at step $t$. $s_{t+1} \sim P(\cdot|s_t,a_t)$ is the subsequent observation. The exploration tree $\mathcal{R}$ represents the collection of trajectories from tasks $\{\tau_i\}_{i=1}^n$, obtained via the recursive exploration process by $M$.

\paragraph{Planning Experience Utilization.}
The agent needs to learn from past explorations and use these experiences to build high-level trajectory data. The coarse high-level tasks have two limitations. (1) The tasks and trajectories are not always aligned. For example, the task requires more than 4.5 stars, but the trajectory only reaches 4 stars. (2) The task lacks stricter constraints for unknown environments, because the websites are partially observable environments. The constraints of the unknown environment must come from real exploration, and the homepage information cannot provide them, such as ingredients and preparation directions. Using such mismatched data causes the agent to learn incorrect patterns and miss key details. Thus, recasting these explorations into accurate experiences is vital for ensuring high-quality training signals.

In the experience extraction stage, the MLLM $M$ compares before-action state and after-action state to extract atomic experiences:
\begin{equation}
\epsilon_t = M(s_t, a_t, s_{t+1}),
\end{equation}
where $s_t$ and $s_{t+1}$ are the visual observations before and after action $a_t$, respectively. A trajectory-level experience $\mu$ can be represented as a sequence of atomic experiences:
\begin{equation}
\mu = (\epsilon_1, \epsilon_2, \ldots, \epsilon_T).
\end{equation}

The agent then fuses these sequences of atomic experiences into refined high-level tasks that are both more aligned with real outcomes and stricter in the constraints. Formally, define a mapping $\Phi$ with $M$ that aggregates the experiences into PEEU task $\tilde{d}$, forming the collection $\tilde{\mathcal{D}}$ of PEEU tasks:
\begin{equation}
\tilde{\mathcal{D}} =  (\tilde{d_1}, \tilde{d_2}, \ldots, \tilde{d_n}) =  \Phi(\mu_1, \mu_2, \ldots, \mu_n, M).
\end{equation}

In the training stage, the agent's goal is to learn a policy $\pi : \mathcal{S} \times \mathcal{H} \times \tilde{\mathcal{D}} \to \mathcal{A}$, that maps the current state $s_t \in \mathcal{S}$, the history $h_t \in \mathcal{H}_{0:t}$, and the task description $\tilde{d} \in \tilde{\mathcal{D}}$, to the next action $a_t \in \mathcal{A}$. We use SFT and GRPO~\citep{shao2024deepseekmath} for training. The details are shown in Appendix~\ref{appendix:PEEU Prompt} and~\ref{appendix:Algorithm Details}.

\subsection{Experimental Settings}

\paragraph{Baseline.}
(1) Atomic-Prompt~\citep{wang2024agent} uses the input task to retrieve related atomic experiences. The number of retrieved atomic experiences is set to 10. These experiences are used as prompts to serve as contextual input.
(2) Trajectory-Prompt~\citep{wang2024agent} uses the input task to retrieve one trajectory-level experience according to its query as the prompt.  
(3) Coarse~\citep{logeswaranscaling,trabucco2025insta} uses the original exploration task as the training task. 
(4) Atomic~\citep{gu2024your,fan2025gui} uses the atomic operation task as the training task.  
In addition, all the training parameters are kept the same. And all methods are controlled to use the same amount of data to ensure a fair comparison.

\paragraph{Evaluation.}
We evaluate the planning capabilities of the models on real-world multimodal benchmark WebVoyager~\citep{he2024webvoyager}. The test set covers diverse real multimodal online websites, including cooking, shopping, research, code, map, study and other categories. Follow the standard evaluation procedure of WebVoyager~\citep{he2024webvoyager}, the benchmark uses the trajectory-level success rate as the final accuracy.

\begin{table*}[!t]
\centering
\setlength{\tabcolsep}{4pt}
\renewcommand{\arraystretch}{1.2}
\resizebox{\textwidth}{!}{
\begin{tabular}{llccccccccc}
\toprule
\multirow{2}{*}{Model} & \multirow{2}{*}{Method} & Allrecipes & Amazon & Apple & Arxiv & Github & Coursera  & Map &  Wolfram & \textbf{Overall} \\
 & & ID & OOD & OOD & OOD & OOD & OOD & OOD & OOD & Total \\
\hline
GPT-4o & \cellcolor{softblue} Vanilla~\citep{hurst2024gpt} & \cellcolor{softblue} 56.3 & \cellcolor{softblue} 53.7 & \cellcolor{softblue} 56.6 & \cellcolor{softblue} 60.5 & \cellcolor{softblue} 57.7  & \cellcolor{softblue} 65.1 &   \cellcolor{softblue} 56.9  & \cellcolor{softblue} 65.2 & \cellcolor{softblue} 59.0 \\
Claude 3 Opus & \cellcolor{softblue} Vanilla~\citep{anthropic2024claude3} & \cellcolor{softblue} 45.9 & \cellcolor{softblue} 58.6 & \cellcolor{softblue} 58.1 & \cellcolor{softblue} 55.0 & \cellcolor{softblue} 56.9 &  \cellcolor{softblue} 68.2 & \cellcolor{softblue} 55.3  & \cellcolor{softblue} 51.5 & \cellcolor{softblue} 56.1 \\
Qwen2.5-VL-72B & \cellcolor{softblue} Vanilla~\citep{bai2025qwen2} & \cellcolor{softblue} 6.6 & \cellcolor{softblue} 58.5 & \cellcolor{softblue} 25.5 & \cellcolor{softblue} 32.5 & \cellcolor{softblue} 17.0 &  \cellcolor{softblue} 21.4 &  \cellcolor{softblue} 36.5 & \cellcolor{softblue} 36.9 &\cellcolor{softblue}  29.3 \\
Qwen2.5-VL-32B & \cellcolor{softblue} Vanilla~\citep{bai2025qwen2} & \cellcolor{softblue} 0.0 & \cellcolor{softblue} 39.0 & \cellcolor{softblue} 16.2 & \cellcolor{softblue} 32.5 & \cellcolor{softblue} 21.9 &  \cellcolor{softblue} 19.0 &  \cellcolor{softblue} 34.1 & \cellcolor{softblue} 19.5 &\cellcolor{softblue}  22.7 \\
\hline
\multirow{9}{*}{\makecell{Qwen2.5-VL-3B\\0.1k trajectories}}
& \cellcolor{warmpink} Vanilla~\citep{bai2025qwen2} & \cellcolor{warmpink} 0.0 & \cellcolor{warmpink} 0.0 & \cellcolor{warmpink} 0.0 & \cellcolor{warmpink} 0.0 & \cellcolor{warmpink} 0.0 & \cellcolor{warmpink} 0.0 & \cellcolor{warmpink} 0.0 & \cellcolor{warmpink} 2.1 & \cellcolor{warmpink} 0.2 \\
& \cellcolor{peach} Atomic-Prompt~\citep{wang2024agent} & \cellcolor{peach} 0.0 & \cellcolor{peach} 0.0 & \cellcolor{peach} 0.0 & \cellcolor{peach} 0.0 & \cellcolor{peach} 0.0 & \cellcolor{peach} 0.0 & \cellcolor{peach} 0.0 & \cellcolor{peach} 0.0 & \cellcolor{peach} 0.0 \textcolor{red!60!black}{\footnotesize (-0.2\%)}\\
 & \cellcolor{peach} Trajectory-Prompt~\citep{wang2024agent} & \cellcolor{peach} 0.0 & \cellcolor{peach} 0.0 & \cellcolor{peach} 0.0 & \cellcolor{peach} 0.0 & \cellcolor{peach} 0.0 & \cellcolor{peach} 0.0 & \cellcolor{peach} 0.0 & \cellcolor{peach} 0.0 & \cellcolor{peach} 0.0 \textcolor{red!60!black}{\footnotesize (-0.2\%)}\\
 & \cellcolor{mintgreen} Coarse-SFT~\citep{logeswaranscaling} & \cellcolor{mintgreen} 0.0 & \cellcolor{mintgreen} 0.0 & \cellcolor{mintgreen} 0.0 & \cellcolor{mintgreen} 2.3 & \cellcolor{mintgreen} 2.4 &  \cellcolor{mintgreen} 4.7 &  \cellcolor{mintgreen} 2.4 & \cellcolor{mintgreen} 4.3 & \cellcolor{mintgreen} 2.0 \textcolor{green!60!black}{\footnotesize (+1.8\%)}\\
 & \cellcolor{mintgreen} Coarse-GRPO~\citep{logeswaranscaling} & \cellcolor{mintgreen} 0.0 & \cellcolor{mintgreen} 2.4 & \cellcolor{mintgreen} 0.0 & \cellcolor{mintgreen} 20.9 & \cellcolor{mintgreen} 0.0 & \cellcolor{mintgreen} 2.3 &  \cellcolor{mintgreen} 2.4 &  \cellcolor{mintgreen} 17.3 & \cellcolor{mintgreen} 5.6 \textcolor{green!60!black}{\footnotesize (+5.4\%)}\\
 & \cellcolor{lightyellow} Atomic-SFT~\citep{fan2025gui} & \cellcolor{lightyellow} 2.2 & \cellcolor{lightyellow} 2.4 & \cellcolor{lightyellow} 0.0 & \cellcolor{lightyellow} 4.6 & \cellcolor{lightyellow} 7.3 & \cellcolor{lightyellow} 7.1 & \cellcolor{lightyellow} 0.0 &  \cellcolor{lightyellow} 15.2 & \cellcolor{lightyellow} 4.8 \textcolor{green!60!black}{\footnotesize (+4.6\%)}\\
 & \cellcolor{lightyellow} Atomic-GRPO~\citep{fan2025gui} & \cellcolor{lightyellow} 0.0 & \cellcolor{lightyellow} 12.1 & \cellcolor{lightyellow} 2.3 & \cellcolor{lightyellow} 11.6 & \cellcolor{lightyellow} 0.0 &  \cellcolor{lightyellow} 9.5 &  \cellcolor{lightyellow} 0.0 &  \cellcolor{lightyellow} 8.6 & \cellcolor{lightyellow} 5.5 \textcolor{green!60!black}{\footnotesize (+5.3\%)}\\
& \cellcolor{lavender} PEEU-SFT (Ours) & \cellcolor{lavender} 2.2 & \cellcolor{lavender} 7.3 & \cellcolor{lavender} 6.9 & \cellcolor{lavender} 11.6 & \cellcolor{lavender} 2.4 & \cellcolor{lavender} 0.0 & \cellcolor{lavender} 4.8 & \cellcolor{lavender}  10.8 & \cellcolor{lavender}  \underline{5.7} \textcolor{green!60!black}{\footnotesize (+5.5\%)}\\
 & \cellcolor{lavender} PEEU-GRPO (Ours) & \cellcolor{lavender} 6.6 & \cellcolor{lavender} 24.3 & \cellcolor{lavender} 3.0 & \cellcolor{lavender} 23.2 & \cellcolor{lavender} 9.7 & \cellcolor{lavender} 7.1 & \cellcolor{lavender} 0.0 & \cellcolor{lavender} 15.2 & \cellcolor{lavender} \textbf{11.1} \textcolor{green!60!black}{\footnotesize (+10.9\%)}\\ 
\hline
\multirow{9}{*}{\makecell{Qwen2.5-VL-7B\\0.1k trajectories}}
& \cellcolor{warmpink} Vanilla~\citep{bai2025qwen2} & \cellcolor{warmpink} 2.2 & \cellcolor{warmpink} 7.3 & \cellcolor{warmpink} 9.3 & \cellcolor{warmpink} 4.6 & \cellcolor{warmpink} 9.7 & \cellcolor{warmpink} 16.6 & \cellcolor{warmpink} 0.0 & \cellcolor{warmpink} 13.0 & \cellcolor{warmpink} 7.8 \\
& \cellcolor{peach} Atomic-Prompt~\citep{wang2024agent} & \cellcolor{peach} 2.2 & \cellcolor{peach} 0.0 & \cellcolor{peach} 6.9 & \cellcolor{peach} 4.6 & \cellcolor{peach} 2.4 & \cellcolor{peach} 9.5 & \cellcolor{peach} 0.0 & \cellcolor{peach} 4.3 & \cellcolor{peach} 3.7 \textcolor{red!60!black}{\footnotesize (-4.1\%)}\\
& \cellcolor{peach} Trajectory-Prompt~\citep{wang2024agent} & \cellcolor{peach} 4.4 & \cellcolor{peach} 0.0 & \cellcolor{peach} 0.0 & \cellcolor{peach} 4.6 & \cellcolor{peach} 2.4 & \cellcolor{peach} 9.5 & \cellcolor{peach} 2.4 & \cellcolor{peach} 6.5 & \cellcolor{peach} 3.7 \textcolor{red!60!black}{\footnotesize (-4.1\%)}\\
& \cellcolor{mintgreen} Coarse-SFT~\citep{logeswaranscaling} & \cellcolor{mintgreen} 0.0 & \cellcolor{mintgreen} 4.8 & \cellcolor{mintgreen} 0.0 & \cellcolor{mintgreen} 4.6 & \cellcolor{mintgreen} 0.0 & \cellcolor{mintgreen} 7.1 & \cellcolor{mintgreen} 4.8 & \cellcolor{mintgreen} 17.3 & \cellcolor{mintgreen} 4.8 \textcolor{red!60!black}{\footnotesize (-3.0\%)} \\
& \cellcolor{mintgreen} Coarse-GRPO~\citep{logeswaranscaling} & \cellcolor{mintgreen} 0.0 & \cellcolor{mintgreen} 17.0 & \cellcolor{mintgreen} 7.1 & \cellcolor{mintgreen} 20.9 & \cellcolor{mintgreen} 4.8 & \cellcolor{mintgreen} 4.7 & \cellcolor{mintgreen} 12.1 & \cellcolor{mintgreen} 26.0 & \cellcolor{mintgreen} 11.5 \textcolor{green!60!black}{\footnotesize (+3.7\%)}\\
& \cellcolor{lightyellow} Atomic-SFT~\citep{fan2025gui} & \cellcolor{lightyellow} 15.5 & \cellcolor{lightyellow} 17.0 & \cellcolor{lightyellow} 11.6 & \cellcolor{lightyellow} 23.2 & \cellcolor{lightyellow} 0.0 & \cellcolor{lightyellow} 7.1 & \cellcolor{lightyellow} 4.8 & \cellcolor{lightyellow} 19.5 & \cellcolor{lightyellow} 12.3 \textcolor{green!60!black}{\footnotesize (+4.5\%)} \\
& \cellcolor{lightyellow} Atomic-GRPO~\citep{fan2025gui} & \cellcolor{lightyellow} 2.2 & \cellcolor{lightyellow} 19.5 & \cellcolor{lightyellow} 0.0 & \cellcolor{lightyellow} 18.6 & \cellcolor{lightyellow} 0.0 & \cellcolor{lightyellow} 11.9 & \cellcolor{lightyellow} 0.0 & \cellcolor{lightyellow} 28.2 & \cellcolor{lightyellow} 10.0 \textcolor{green!60!black}{\footnotesize (+2.2\%)} \\
& \cellcolor{lavender} PEEU-SFT (Ours) & \cellcolor{lavender} 8.8 & \cellcolor{lavender} 24.3 & \cellcolor{lavender} 18.6 & \cellcolor{lavender} 16.2 & \cellcolor{lavender} 7.3 & \cellcolor{lavender} 16.6 & \cellcolor{lavender} 7.3 & \cellcolor{lavender} 26.0 & \cellcolor{lavender} \underline{15.6} \textcolor{green!60!black}{\footnotesize (+7.8\%)} \\
& \cellcolor{lavender} PEEU-GRPO (Ours) & \cellcolor{lavender} 4.4 & \cellcolor{lavender} 26.8 & \cellcolor{lavender} 18.6 & \cellcolor{lavender} 20.9 & \cellcolor{lavender} 21.9 & \cellcolor{lavender} 33.3 & \cellcolor{lavender} 12.1 & \cellcolor{lavender} 21.7 & \cellcolor{lavender}  \textbf{19.9} \textcolor{green!60!black}{\footnotesize (+12.1\%)} \\
\hline
\multirow{4}{*}{\makecell{Qwen2.5-VL-3B\\2k trajectories}}
& \cellcolor{warmpink} Vanilla~\citep{bai2025qwen2} & \cellcolor{warmpink} 0.0 & \cellcolor{warmpink} 0.0 & \cellcolor{warmpink} 0.0 & \cellcolor{warmpink} 0.0 & \cellcolor{warmpink} 0.0 & \cellcolor{warmpink} 0.0 & \cellcolor{warmpink} 0.0 & \cellcolor{warmpink} 2.1 & \cellcolor{warmpink} 0.2  \\
& \cellcolor{mintgreen} Coarse-SFT~\citep{logeswaranscaling} & \cellcolor{mintgreen} 0.0 & \cellcolor{mintgreen} 12.1 & \cellcolor{mintgreen} 6.9 & \cellcolor{mintgreen} 6.9 & \cellcolor{mintgreen} 9.7 & \cellcolor{mintgreen} 14.2 &  \cellcolor{mintgreen} 17.0 & \cellcolor{mintgreen} 39.1 & \cellcolor{mintgreen} 13.2 \textcolor{green!60!black}{\footnotesize (+13.0\%)} \\
& \cellcolor{lightyellow} Atomic-SFT~\citep{fan2025gui} & \cellcolor{lightyellow} 13.3 & \cellcolor{lightyellow} 26.8 & \cellcolor{lightyellow} 4.6 & \cellcolor{lightyellow} 23.2 & \cellcolor{lightyellow} 7.3 &  \cellcolor{lightyellow} 9.5  & \cellcolor{lightyellow} 17.0 & \cellcolor{lightyellow} 32.6 & \cellcolor{lightyellow} 16.7 \textcolor{green!60!black}{\footnotesize (+16.5\%)} \\
& \cellcolor{lavender} PEEU-SFT (Ours) & \cellcolor{lavender} 8.8 & \cellcolor{lavender} 46.3 & \cellcolor{lavender} 13.9 & \cellcolor{lavender} 13.9 & \cellcolor{lavender} 9.7 & \cellcolor{lavender} 14.2 & \cellcolor{lavender} 21.9 &  \cellcolor{lavender} 30.4 & \cellcolor{lavender} \textbf{19.8} \textcolor{green!60!black}{\footnotesize (+19.6\%)} \\
\hline
\multirow{4}{*}{\makecell{Qwen2.5-VL-7B\\2k trajectories}}
& \cellcolor{warmpink} Vanilla~\citep{bai2025qwen2} & \cellcolor{warmpink} 2.2 & \cellcolor{warmpink} 7.3 & \cellcolor{warmpink} 9.3 & \cellcolor{warmpink} 4.6 & \cellcolor{warmpink} 9.7 & \cellcolor{warmpink} 16.6 & \cellcolor{warmpink} 0.0 & \cellcolor{warmpink} 13.0 & \cellcolor{warmpink} 7.8 \\
& \cellcolor{mintgreen} Coarse-SFT~\citep{logeswaranscaling} & \cellcolor{mintgreen} 6.6 & \cellcolor{mintgreen} 34.1 & \cellcolor{mintgreen} 20.9 & \cellcolor{mintgreen} 20.9 & \cellcolor{mintgreen} 17.0 & \cellcolor{mintgreen} 14.2 &  \cellcolor{mintgreen} 17.0 & \cellcolor{mintgreen} 21.7 & \cellcolor{mintgreen} 19.0 \textcolor{green!60!black}{\footnotesize (+11.2\%)} \\
& \cellcolor{lightyellow} Atomic-SFT~\citep{fan2025gui} & \cellcolor{lightyellow} 13.3 & \cellcolor{lightyellow} 51.2 & \cellcolor{lightyellow} 6.9 & \cellcolor{lightyellow} 25.5 & \cellcolor{lightyellow} 0.0 &  \cellcolor{lightyellow} 9.5  & \cellcolor{lightyellow} 39.0 & \cellcolor{lightyellow} 28.2 & \cellcolor{lightyellow} 21.7 \textcolor{green!60!black}{\footnotesize (+13.9\%)} \\
& \cellcolor{lavender} PEEU-SFT (Ours) & \cellcolor{lavender} 17.7 & \cellcolor{lavender} 53.6 & \cellcolor{lavender} 16.2 & \cellcolor{lavender} 25.2 & \cellcolor{lavender} 19.5 & \cellcolor{lavender} 35.7 & \cellcolor{lavender} 48.7 &  \cellcolor{lavender} 28.2 & \cellcolor{lavender} \textbf{30.6} \textcolor{green!60!black}{\footnotesize (+22.8\%)} \\
\bottomrule
\end{tabular}
}
\caption{Performance across different OOD websites. Bold indicates the highest performance. Underline indicates the second-highest performance. Overall is the average accuracy of all websites.}
\label{tab:agent_results}

\end{table*}

\paragraph{Exploration and training settings.}
%对于探索阶段，使用GPT-4o进行探索，最大rollout长度为15，探索轨迹数量为100，浏览器观测的像素为1024*768。对于经验总结阶段，使用GPT-4o对浏览器前后状态的变化进行经验总结。
%对于注入阶段
%对于训练和测试的划分，为了充分验证模型的通用和泛化能力，我们只在Allrecipes一个网站上训练，剩余的14个网站在训练期间未见过。
(1) For the exploration phase, we use GPT-4o for exploration with a maximum step length of 15 in 0.1k or 2k exploration tasks. For the experience summarization phase, we use GPT-4o to summarize the changes in the browser's state before and after the exploration. %The browser observation resolution is set to 1024*768 pixels.
(2) For the training phase, all our experiments are conducted on Qwen2.5-VL-3B-Instruct and Qwen2.5-VL-7B-Instruct. For the SFT model, the batch size is 16, the learning rate is 5.0e-6, and the number of training epochs is 5, using the llama-factory~\citep{zheng2024llamafactory} training framework. For the GRPO model, the batch size is 20, the learning rate is 1.0e-6, the rollout size is 10, and the number of training epochs is 7, using the verl~\citep{zheng2025easyr1} framework. All experiments are performed on 4 A800 GPUs. For fair comparison, all experiments use identical trajectory scales.
(3) Our experimental setup consists of two configurations: the first involves training on 0.1k trajectories derived from Allrecipes, while the second utilizes 2k trajectories from a previously unseen website. We test on seven additional websites that were entirely excluded from the training data. More details are shown in Appendix~\ref{appendix:PEEU Dataset Details}.

\subsection{Results and Analysis}

\paragraph{Adapt the task to fit the trajectory with experience.}
As shown in Figure~\ref{fig:Method}, coarse trajectory tasks face problems of mismatch and a lack of strict constraints. For example, in the coarse task, the rating is 4.5, but the trajectory shows only 4 stars, which causes a mismatch.  Therefore, constraints should be derived from exploration experience. By using experience to modify tasks, we can create more aligned and strictly constrained advanced tasks. As shown in Table~\ref{tab:agent_results}, for the 7B model, trained with 2k trajectories, PEEU-SFT achieves a remarkable overall score of 30.6\%, which not only significantly outperforms the competitive Coarse-SFT of 19.0\%  baselines but also surpasses the much larger Qwen2.5-VL-32B Instruct model of 22.7\%. This underscores the efficiency of deriving strict constraints from exploration experience to enhance model capability. Furthermore, our method consistently demonstrates superior performance and substantial gains across varying model scales (3B and 7B) and data quantities (0.1k and 2k), validating the general effectiveness of PEEU in diverse settings.

\paragraph{Using higher-level tasks provides better cross-website generalization than lower-level tasks in real-world websites.}
As illustrated in Table~\ref{tab:agent_results}, relying on atomic-level tasks, limits cross-website generalization. Therefore, higher-level tasks like PEEU are essential for enhancing the generalization capability of task decomposition across different websites. For the Qwen2.5-VL-7B model trained with 2k trajectories, our PEEU-SFT achieves a overall accuracy of 30.6\%, outperforming the Atomic-SFT baseline of 21.7\%. For the Qwen2.5-VL-7B model trained with 0.1k trajectories, our PEEU-GRPO achieves a overall accuracy of 19.9\%, outperforming the Atomic-GRPO baseline of 10.0\%. Furthermore, this trend is consistent across different model sizes (e.g., Qwen2.5-VL-3B and Qwen2.5-VL-7B) and data regimes (e.g., 0.1k and 2k trajectories), showing that higher-level tasks provide a more generalization ability for planning in the unseen web environments.

\paragraph{Without a specially designed prompt pipeline, direct training is more effective than retrieval for small models.}
As shown in Table~\ref{tab:agent_results}, we apply both training and retrieval under the same experiences. Because of the limited ability of small models, using prompts without changing model parameters does not effectively help them improve in complex tasks. For example, with the retrieval method, a 7B model gets scores of 3.7\% for both Atomic-Prompt and Trajectory-Prompt, which are even lower than the base model score of 7.8\% because their reasoning capabilities are too limited without any training. Similarly, the 3B model fails to effectively utilize retrieved context, resulting in 0.0\% accuracy across prompt-based methods. In stark contrast, training methods yield substantial gains; specifically, PEEU-GRPO boosts the 7B and 3B models to 19.9\% and 11.1\% respectively. This shows direct training is more effective than retrieval for small models.

\begin{figure*}[h]
\begin{center}
\includegraphics[width=\linewidth]{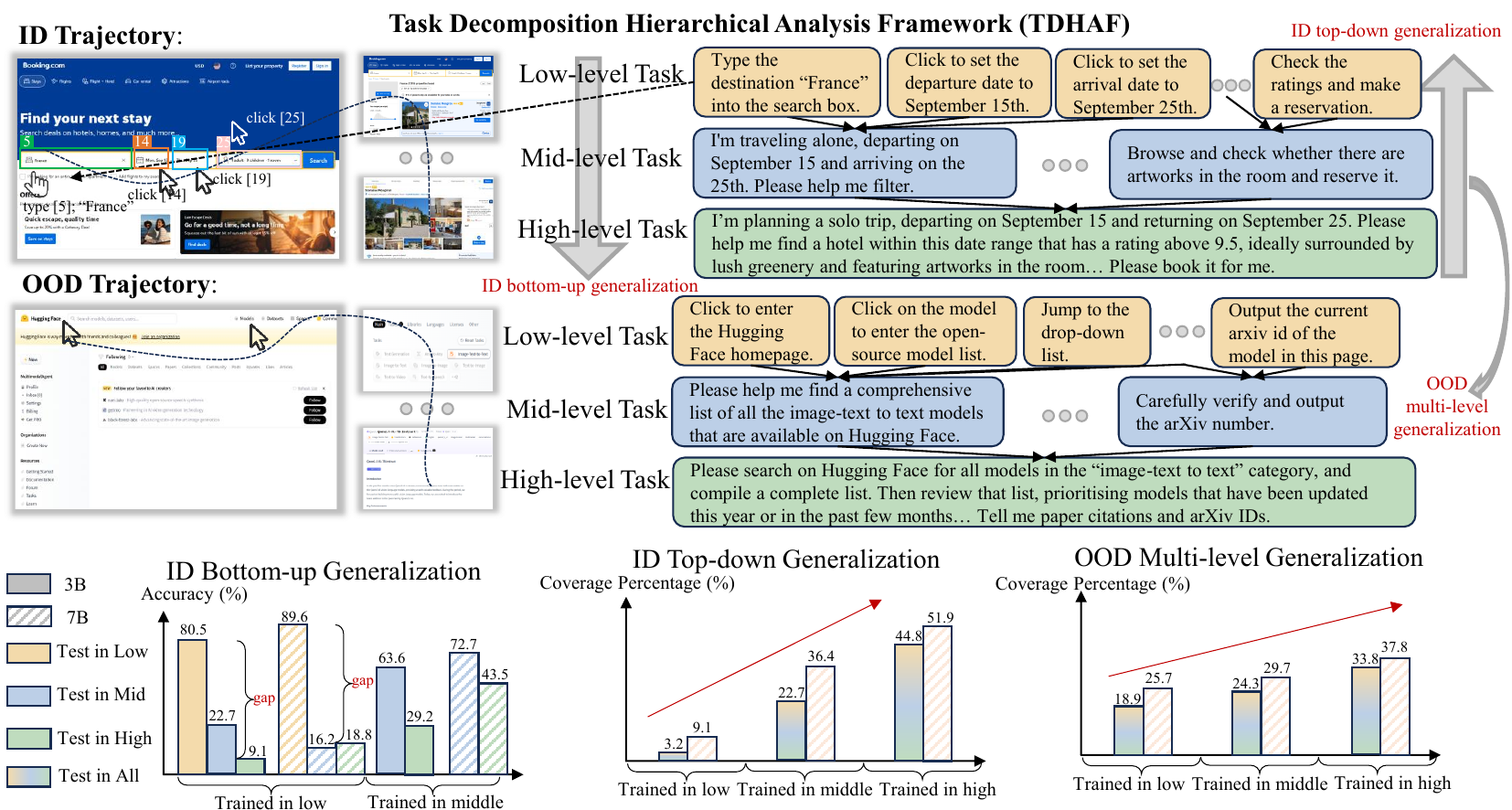}
\end{center}
\caption{This figure illustrates the task decomposition hierarchical analysis framework. The upper part shows the trajectory of ID, and the lower part shows the trajectory of OOD. Both domains contain three levels: low, middle, and high. We study three generalization dimensions, including ID bottom-up generalization, ID top-down generalization and OOD multi-level generalization.}
\label{fig:framework}
\end{figure*}

\section{Task Decomposition Hierarchical Analysis Framework}

Although PEEU achieves stronger performance, a critical research question remains: what is the agent's capacity for compositional generalization across different levels of task decomposition? To answer this and analyze the hierarchical generalization capabilities of task decomposition, we propose the \textbf{Task Decomposition Hierarchical Analysis Framework} (\textbf{TDHAF}). Illustrated in Figure~\ref{fig:framework}, this framework enables a rigorous evaluation from three perspectives: \textbf{ID bottom-up generalization}, \textbf{ID top-down generalization}, and \textbf{OOD multi-level generalization} (Appendix~\ref{appendix:Definition Details}). In this section, we introduce the analysis framework, data construction, experimental settings, results and analysis.
\subsection{Analysis Framework}
To investigate the compositional generalization ability of models in multimodal web navigation task planning scenarios, we propose the task decomposition hierarchical analysis framework. This framework first defines three levels of task granularity: \textbf{low-level} tasks, \textbf{mid-level} tasks, and \textbf{high-level} tasks. It further distinguishes between two types of generalization: in-domain (\textbf{ID}) and out-of-domain (\textbf{OOD}). Building on this taxonomy, the framework analyzes from three perspectives: bottom-up generalization in-domain, top-down generalization in-domain, and multi-level generalization out-of-domain. Figure~\ref{fig:framework} provides a detailed example of the analysis framework. Table~\ref{tab:generalization_dimensions} illustrates the training and testing set divisions for the three generalization dimensions. Explanations of the three dimensions of generalization are presented following.

\begin{table*}[!t]
\centering
\caption{Accuracy comparison across different generalization dimensions. 3B Instruct refers to the Qwen2.5-VL-3B-Instruct model. 3B Low refers to the Qwen2.5-VL-3B-Instruct trained at the low level (atomic level). 3B High refers to the Qwen2.5-VL-3B-Instruct trained at the high level. Test-ID-Low denotes the in-domain low-level test set. Test-OOD-Low denotes the out-of-domain low-level test set. The bolded entries indicate the model that achieves the highest Step SR among the four models on each test set under the same base model.}
\resizebox{\textwidth}{!}{ % Adjusts table width to fit the page
\begin{tabular}{lcccccccccccc}
\toprule
\multirow{2}{*}{Model} & \multicolumn{4}{c}{Test-ID-Low} & \multicolumn{4}{c}{Test-ID-Middle} & \multicolumn{4}{c}{Test-ID-High} \\
\cmidrule(lr){2-5} \cmidrule(lr){6-9} \cmidrule(lr){10-13}
 & Id & Action & Value & \textbf{Step SR} & Id & Action & Value & \textbf{Step SR} & Id & Action & Value & \textbf{Step SR} \\
\midrule
3B Instruct & 30.3 & 39.5 & 85.7 & 17.8 & 17.1 & 6.6 & 9.5 & 0.0 & 14.4 & 9.6 & 6.7 & 0.7 \\
3B Low & 81.2 & 99.4 & 100.0 & \textbf{80.5} & 28.6 & 83.1 & 4.3 & 22.7 & 12.3 & 85.1 & 0.0 & 9.1 \\
3B Middle & 72.7 & 98.7 & 95.7 & 71.4 & 66.9 & 95.5 & 73.9 & \textbf{63.6} & 32.5 & 85.1 & 0.0 & 29.2 \\
3B High & 77.3 & 98.1 & 95.7 & 75.3 & 57.8 & 94.2 & 65.2 & 54.5 & 64.9 & 95.5 & 65.2 & \textbf{63.0} \\
\midrule
7B Instruct & 59.1 & 84.4 & 73.9  & 49.4 & 43.1 & 41.2 & 27.3 & 17.6 & 35.8 & 44.4 & 20.0 & 13.2 \\
7B Low & 90.3 & 99.4 & 100.0 & \textbf{89.6} & 37.7 & 39.6 & 43.5 & 16.2 & 29.2 & 75.3 & 13.0 & 18.8 \\
7B Middle & 87.0 & 99.4 & 95.7 & 86.4 & 78.6 & 92.2 & 65.2 & \textbf{72.7} & 46.1 & 89.6 & 21.7 & 43.5 \\
7B High & 85.1 & 98.1 & 87.0 & 83.1 & 69.5 & 89.6 & 39.1 & 63.6 & 76.6 & 92.2 & 56.5 & \textbf{72.1} \\
\midrule
\multirow{2}{*}{Model} & \multicolumn{4}{c}{Test-OOD-Low} & \multicolumn{4}{c}{Test-OOD-Middle} & \multicolumn{4}{c}{Test-OOD-High} \\
\cmidrule(lr){2-5} \cmidrule(lr){6-9} \cmidrule(lr){10-13}
 & Id & Action & Value & \textbf{Step SR} & Id & Action & Value & \textbf{Step SR} & Id & Action & Value & \textbf{Step SR} \\
\midrule
3B Instruct & 40.5 & 63.5 & 100.0 & 31.1 & 21.9 & 20.5 & 33.3 & 6.8 & 16.4 & 16.4 & 22.2 & 0.0 \\
3B Low & 81.1 & 98.6 & 100.0 & 79.7 & 37.8 & 75.7 & 12.5 & 35.1 & 29.7 & 78.4 & 0.0 & 25.7 \\
3B Middle & 70.3 & 100.0 & 100.0 & 70.3 & 48.6 & 79.7 & 12.5 & \textbf{44.6} & 32.4 & 78.4 & 0.0 & 31.1 \\
3B High & 82.4 & 100.0 & 100.0 & \textbf{82.4} & 45.9 & 81.1 & 12.5 & \textbf{44.6} & 39.2 & 81.1 & 6.2 & \textbf{39.2} \\
\midrule
7B Instruct & 63.5 & 91.9 & 62.5 & 56.8 & 46.6 & 72.6 & 20.0 & 30.1 & 30.1 & 64.4 & 20.0 & 16.4\\
7B Low & 89.2 & 97.3 & 93.8 & \textbf{85.1} & 56.8 & 78.4 & 31.2 & 50.0 & 37.8 & 79.7 & 18.8 & 33.8 \\
7B Middle & 83.8 & 100.0 & 93.8 & 82.4 & 59.5 & 82.4 & 12.5 & 51.4 & 37.8 & 78.4 & 0.0 & 35.1 \\
7B High & 81.1 & 95.9 & 75.0 & 77.0 & 58.1 & 82.4 & 12.5 & \textbf{54.1} & 45.9 & 81.1 & 6.2 & \textbf{43.2} \\
\bottomrule
\end{tabular}
}
\label{tab:accuracy}
\vspace{-15pt}
\end{table*}

\paragraph{ID Bottom-up Generalization.}
To study whether the model can generalize from low-level tasks to higher-level composite tasks in-domain, we use relatively low-level tasks as the training set and high-level tasks as the test set. For example, after the model learns single-step atomic task mapping, we test if it can generalize to multi-step subtasks and long-horizon task decomposition. We test if it can generalize to long-horizon task decomposition after learning subtasks.

\paragraph{ID Top-down Generalization.} 
To study whether the model can generalize from high-level tasks to lower-level tasks in-domain, we use relatively high-level tasks as the training set and relatively low-level tasks as the test set, which is the opposite of the previous experiment. For example, after the model learns to decompose long-horizon tasks, we check whether it truly learns the corresponding subtasks and atomic skills.

\paragraph{OOD Multi-level Generalization.} 
To study whether the model can generalize task decomposition ability from in-domain tasks to out-of-domain tasks, we separately use three levels of in-domain tasks as the training set. We use unseen cross-website tasks as the test set to evaluate multi-level out-of-domain generalization. For example, we examine how well it applies abilities to unseen tasks.

\begin{figure*}[t]
\begin{center}
\includegraphics[width=\linewidth]{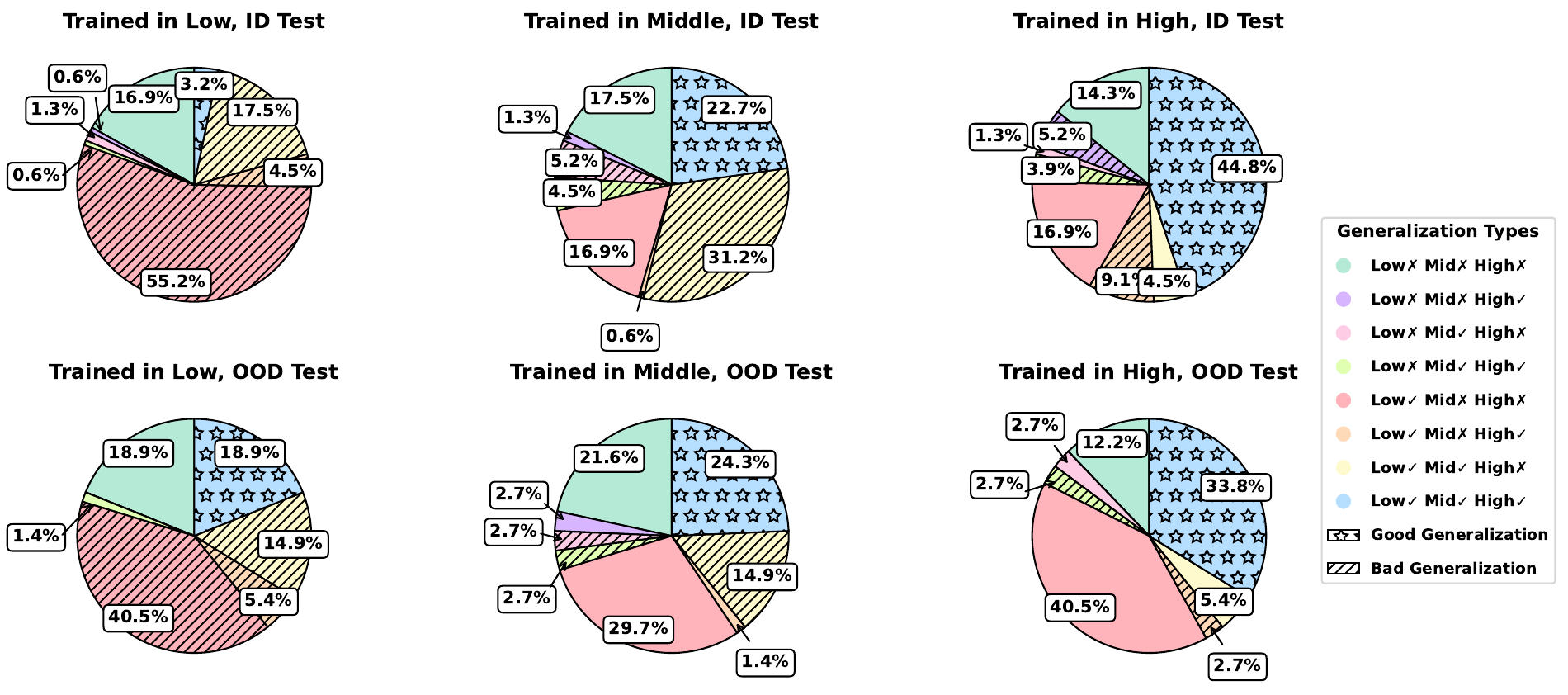}
\end{center}
\caption{Generalization distribution pie chart for Qwen2.5-VL-3B. The table shows the distribution of eight types of generalization. Good generalization means successful generalization to other levels, the larger the better. The good generalization area expands from left to right, demonstrating high-level task training yields better generalization. Results for Qwen2.5-VL-7B. The definitions of good/bad generalization are shown in Appendix~\ref{appendix:Generalization Distribution and Definition}.}
\label{fig:analysis_pie_3B}
\vspace{-10pt}
\end{figure*}

\subsection{Experimental Settings}
\paragraph{Settings.} 
All experiments are conducted on Qwen2.5-VL-3B-Instruct and Qwen2.5-VL-7B-Instruct for SFT. The batch size is 8, the learning rate is 5.0e-6 and the training epochs are 3, with llama-factory~\citep{zheng2024llamafactory} framework. All experiments are conducted on 4 A800 GPUs.
\paragraph{Metric.} 
Following~\citep{deng2023mindweb,zheng2024seeact}, we calculate the accuracy between predictions and ground truth, which includes the following four sub-metrics: \textit{Id} refers to the accuracy of interactive element number in the Set-of-Mark (SoM). \textit{Action} measures the accuracy of action types. \textit{Value} evaluates the accuracy of action parameters. \textit{Step SR} represents the accuracy rate of a single-step prediction completely matching the ground truth.

\subsection{Results and Analysis}

\paragraph{Mastering individual low-level tasks does not necessarily imply mastery of the corresponding high-level task.}
As shown in Table~\ref{tab:accuracy} and Figure~\ref{fig:framework} in the Step SR in-domain setting, the 3B-model trained in low-level training data achieves 80.5\% accuracy in low-level test tasks, but only 9.1\% accuracy for the corresponding high-level test tasks. Similarly, the 7B-model trained on low-level data achieves 89.6\% accuracy on low-level tasks, but only 18.8\% on high-level ones. This shows that the bottom-up post-training method is not an effective way for enhancing planning ability.

\paragraph{Using high-level tasks makes it easier to generalize downwards in-domain with greater overall coverage.}
As shown in Figure~\ref{fig:analysis_pie_3B} and Figure~\ref{fig:analysis_pie_7B} in the in-domain setting, we define a task where all levels succeed as good generalization, and we refer to this percentage as the coverage percentage (Appendix~\ref{appendix:Generalization Distribution and Definition} for a formal definition). For the 3B model, the coverage percentage is 44.8\% when trained on high-level tasks, 22.7\% on middle-level, and 3.2\% on low-level tasks. The 7B model achieves 51.9\% (high-level), 36.4\% (middle-level), and 9.1\% (low-level) coverage. This shows top-down generalization has higher coverage percentage in-domain.

\paragraph{Using high-level task training can enable the model to acquire stronger generalization capabilities for multi-level tasks in OOD.}
As shown in Figure~\ref{fig:analysis_pie_3B} and Figure~\ref{fig:analysis_pie_7B} in the out-of-domain setting, for the 3B model, the coverage percentage is 33.8\% when trained on high-level tasks, 24.3\% on middle-level, and 18.9\% on low-level tasks. For the 7B model, the coverage percentage is 37.8\% when trained on high-level tasks, 29.7\% on middle-level, and 25.7\% on low-level tasks. This shows that top-down generalization also has higher coverage percentage out-of-domain.

\section{Related Work}

\paragraph{DeepResearch Agent.}
DeepResearch emphasizes broad web searches~\citep{zhang2025web,li2025towards}. Systems like WebSailor~\citep{li2025websailor}, WebShaper~\citep{tao2025webshaper}, and WebWatcher~\citep{geng2025webwatcher} focus on information seeking. But experience summarization and compositional generalization analysis~\citep{li2025unveiling} remain underexplored. AWM~\citep{wang2024agent}, Agent KB~\citep{tang2025agent}, Memento~\citep{zhou2025memento} and Memp~\citep{fang2025memp} construct structured knowledge bases from past explorations using prompt engineering without training. To bridge this gap, we study compositional generalization in task planning and leverage automatically mined experiences to train agents, enabling them to achieve stronger web‑based planning capabilities under the same scale of data.

\paragraph{Multimodal Web Navigation Agent.}
The research on multimodal web agent navigation emphasizes vertical depth navigation on web pages~\citep{wang2024gui,ning2025survey,zhou2025proposer,tang2025survey,li2026agentic}. Open-source models need two core abilities: grounding and planning~\citep{wang2024gui,men2024unlocking,nguyen2025gui}. Some works strengthen grounding for more accurate spatial coordinates~\citep{lu2025ui,luo2025gui,zhou2505gui}. The SoM representation can reduce the influence of grounding, making it easier to study improvements in planning ability. Prior work often trains on low-level tasks~\citep{gu2024your,fan2025gui} or distills teacher trajectories without fully utilizing experiences~\citep{logeswaranscaling,trabucco2025insta}. Some works equip agents with memory to enhance their planning capabilities~\citep{hu2025memory,wang2024agent,men2025troublemaker,xia2026skillrl}. Additionally, some studies aim to improve trajectory quality. One group of methods uses a reward model to filter trajectories~\citep{men2025agent,lin2025cuarewardbench,jin2025rag}, and another gives the model the ability to adapt to its environment with the trajectories~\citep{su2025learn,zhou2025proposer,sun2025genesis}. Our approach makes high-level tasks more aligned and constrained, and by leveraging the TDHAF framework to quantitatively analyze this capacity from the perspective of planning granularity, thereby providing stronger generalization ability in the same data scale setting.

\section{Conclusion}
In this work, we propose the Planning Experience Exploration and Utilization (PEEU) method to enhance small MLLMs by leveraging autonomous exploration and hindsight experience. To analyze this, we introduce the Task Decomposition Hierarchical Analysis Framework (TDHAF) to systematically evaluate planning compositional generalization. Experiments show PEEU significantly outperforms larger models on OOD websites, demonstrating training on aligned high-level tasks is effective for planning ability generalization.

\section*{Limitations}
Currently, our evaluation focuses on information-seeking and navigation tasks across diverse real-world websites. Due to privacy and security constraints, we did not include scenarios involving sensitive operations such as user login, CAPTCHA solving, or actual payment transactions. While the proposed high-level task planning is theoretically applicable to these scenarios, extending the agent's capabilities to handle authenticated sessions and security protocols remains a direction for future research.

\section*{Acknowledgements}
This work was supported by the National Natural Science Foundation of China (No.U24A20335, No.62406321), Beijing Natural Science Foundation (L243006), and the independent research project of the Key Laboratory of Cognition and Decision Intelligence for Complex Systems.

\bibliography{main}

\appendix

\newpage

\section{PEEU Prompt}
\label{appendix:PEEU Prompt}

\begin{dialogbox}
Inference Prompt for WebVoyager

System:
Imagine you are a robot browsing the web, just like humans. Now you need to complete a task. In each iteration, you will receive an Observation that includes a screenshot of a webpage and some texts. This screenshot will feature Numerical Labels placed in the TOP LEFT corner of each Web Element.
Carefully analyze the visual information to identify the Numerical Label corresponding to the Web Element that requires interaction, then follow the guidelines and choose one of the following actions:
1. Click a Web Element.
2. Delete existing content in a textbox and then type content. 
3. Scroll up or down. Multiple scrolls are allowed to browse the webpage. Pay attention!! The default scroll is the whole window. If the scroll widget is located in a certain area of the webpage, then you have to specify a Web Element in that area. I would hover the mouse there and then scroll.
4. Wait. Typically used to wait for unfinished webpage processes, with a duration of 5 seconds.
5. Go back, returning to the previous webpage.
6. Google, directly jump to the Google search page. When you can't find information in some websites, try starting over with Google.
7. Answer. This action should only be chosen when all questions in the task have been solved.

Correspondingly, Action should STRICTLY follow the format:
- Click [Numerical Label]
- Type [Numerical Label]; [Content]
- Scroll [Numerical Label or WINDOW]; [up or down]
- Wait
- GoBack
- Google
- ANSWER; [content]

Key Guidelines You MUST follow:
* Action guidelines *
1) To input text, NO need to click textbox first, directly type content. After typing, the system automatically hits `ENTER` key. Sometimes you should click the search button to apply search filters. Try to use simple language when searching.  
2) You must Distinguish between textbox and search button, don't type content into the button! If no textbox is found, you may need to click the search button first before the textbox is displayed. 
3) Execute only one action per iteration. 
4) STRICTLY Avoid repeating the same action if the webpage remains unchanged. You may have selected the wrong web element or numerical label. Continuous use of the Wait is also NOT allowed.
5) When a complex Task involves multiple questions or steps, select "ANSWER" only at the very end, after addressing all of these questions (steps). Flexibly combine your own abilities with the information in the web page. Double check the formatting requirements in the task when ANSWER. 
* Web Browsing Guidelines *
1) Don't interact with useless web elements like Login, Sign-in, donation that appear in Webpages. Pay attention to Key Web Elements like search textbox and menu.
2) Vsit video websites like YouTube is allowed BUT you can't play videos. Clicking to download PDF is allowed and will be analyzed by the Assistant API.
3) Focus on the numerical labels in the TOP LEFT corner of each rectangle (element). Ensure you don't mix them up with other numbers (e.g. Calendar) on the page.
4) Focus on the date in task, you must look for results that match the date. It may be necessary to find the correct year, month and day at calendar.
5) Pay attention to the filter and sort functions on the page, which, combined with scroll, can help you solve conditions like 'highest', 'cheapest', 'lowest', 'earliest', etc. Try your best to find the answer that best fits the task.

For example:
Click [3]
Type [3]; [apple]
Scroll [WINDOW]; [down]
Wait
GoBack
Google
ANSWER; [apple is red]

Your reply should strictly follow the format:
Thought: {Your brief thoughts (briefly summarize the info that will help ANSWER)}
Action: {One Action format you choose}

Then the User will provide:
Observation: {A labeled screenshot Given by User}

User:
<image>Now given a task: <task> Please interact with https://www.example.com and get the answer. Observation: please analyze the attached screenshot and give the Thought and Action. I've provided the tag name of each element and the text it contains (if text exists). Note that <textarea> or <input> may be textbox, but not exactly. Please focus more on the screenshot and then refer to the textual information. <SoM Observation>
\end{dialogbox}

\begin{dialogbox}
Task Setting Prompt

<image>
Analyze the given webpage screenshot and generate 50 different tasks that users might want to accomplish on this website. 
You can focus on searching for specific items. The task should be combined with the specific function of this website.
The tasks should be varied, and there should be both difficult and simple tasks.
Output only a JSON-formatted list of tasks with no additional commentary or explanation.
Example format:
{
"tasks": [
"task 1 description",
"task 2 description",
...
"task n description"
]
}
\end{dialogbox}

\begin{dialogbox}
Exploration Prompt

Imagine you are a robot browsing the web, just like humans. Now you need to complete a task. In each iteration, you will receive an Observation that includes a screenshot of a webpage and some texts. This screenshot will feature Numerical Labels placed in the TOP LEFT corner of each Web Element.
Carefully analyze the visual information to identify the Numerical Label corresponding to the Web Element that requires interaction, then follow the guidelines and choose one of the following actions:
1. Click a Web Element.
2. Delete existing content in a textbox and then type content. 
3. Scroll up or down. Multiple scrolls are allowed to browse the webpage. Pay attention!! The default scroll is the whole window. If the scroll widget is located in a certain area of the webpage, then you have to specify a Web Element in that area. I would hover the mouse there and then scroll.
4. Wait. Typically used to wait for unfinished webpage processes, with a duration of 5 seconds.
5. Go back, returning to the previous webpage. If you scroll down more than twice and still can't find the answer, you need to use "Go back" to return.
6. Google, directly jump to the Google search page. When you can't find information in some websites, try starting over with Google.
7. Answer. This action should only be chosen when all questions in the task have been solved.

Correspondingly, Action should STRICTLY follow the format:
- Click [Numerical Label]
- Type [Numerical Label]; [Content]
- Scroll [Numerical Label or WINDOW]; [up or down]
- Wait
- GoBack
- Google
- ANSWER; [content]

Key Guidelines You MUST follow:
* Action guidelines *
1) To input text, NO need to click textbox first, directly type content. After typing, the system automatically hits `ENTER` key. Sometimes you should click the search button to apply search filters. Try to use simple language when searching.  
2) You must Distinguish between textbox and search button, don't type content into the button! If no textbox is found, you may need to click the search button first before the textbox is displayed. 
3) Execute only one action per iteration. 
4) STRICTLY Avoid repeating the same action if the webpage remains unchanged. You may have selected the wrong web element or numerical label. Continuous use of the Wait is also NOT allowed.
5) When a complex Task involves multiple questions or steps, select "ANSWER" only at the very end, after addressing all of these questions (steps). Flexibly combine your own abilities with the information in the web page. Double check the formatting requirements in the task when ANSWER. 
6) If you feel the current product does not meet the task requirements, you can use GoBack action to return to the previous screen and look for other products. Don't just scroll down-learn to go back.
* Web Browsing Guidelines *
1) Don't interact with useless web elements like Login, Sign-in, donation that appear in Webpages. Pay attention to Key Web Elements like search textbox and menu.
2) Vsit video websites like YouTube is allowed BUT you can't play videos. Clicking to download PDF is allowed and will be analyzed by the Assistant API.
3) Focus on the numerical labels in the TOP LEFT corner of each rectangle (element). Ensure you don't mix them up with other numbers (e.g. Calendar) on the page.
4) Focus on the date in task, you must look for results that match the date. It may be necessary to find the correct year, month and day at calendar.
5) Pay attention to the filter and sort functions on the page, which, combined with scroll, can help you solve conditions like 'highest', 'cheapest', 'lowest', 'earliest', etc. Try your best to find the answer that best fits the task.

Your reply should strictly follow the format:
Thought: {Your brief thoughts (briefly summarize the info that will help ANSWER)}
Action: {One Action format you choose}

Then the User will provide:
Observation: {A labeled screenshot Given by User}
\end{dialogbox}

\begin{dialogbox}

Experience Extraction Prompt

Analyze the user's intent based on the following:
The action performed between these interfaces is <ACTION>

Task: 
The first screenshot shows the interface before interaction, while the second screenshot displays the interface after the click operation.
Generate descriptions explaining the purpose of interaction with the element. 
Focus on meaningful UI changes (e.g., new elements, transitions, or data updates, Don't pay attention to the changes in the bbox.).
Only output the task descriptions experience.
\end{dialogbox}

\begin{dialogbox}

Experience Aggregation Prompt

In this task, there are too many details provided. I only want to keep the details specified by the user, and the specific operational details need to be deleted.
Please directly output the processed string.
The task requirement is a declarative sentence, appearing like a real world user task.

The raw task is as follows:<low-level task list>
\end{dialogbox}

\section{PEEU Algorithm Details}
\label{appendix:Algorithm Details}
The PEEU algorithm is shown in Algorithm~\ref{alg:method}.

\begin{algorithm}[h]
\caption{Autonomous Planning with Exploration and Experience Utilization}
\label{alg:method}
\begin{algorithmic}[1]
\Require Website URL, MLLM $M$, Environment $\text{Env}$
\Ensure  Policy $\pi$ for task-oriented planning

\Statex \textbf{Stage 1: Planning Tree Exploration}
\State Obtain homepage state $s_0$ from the given URL
\State Generate task list: $\mathcal{D} = M(s_0, \text{URL})$
\For{each task $d_i \in \mathcal{D}$}
    \State Execute actions $a_t$ guided by $M$
    \State Transition: $s_{t+1} \sim P(\cdot|s_t,a_t)$
    \State Record trajectory $\tau = (s_0,a_0,s_1,a_1,\dots)$
\EndFor
\State Build exploration tree $\mathcal{R} = \text{Explore}(M,\mathcal{T},\text{Env},\text{URL})$

\Statex \textbf{Stage 2: Planning Experience Utilization}
\For{each trajectory $\tau$}
    \State Extract atomic experiences $\epsilon_t = (s_t, a_t, s_{t+1})$
    \State Build $\mu = (\epsilon_0,\epsilon_1, \ldots, \epsilon_T)$
    \State Fuse into PEEU task: $\tilde{d} = \Phi(\mu)$
\EndFor
\State Train policy $\pi$ with SFT and GRPO using PEEU dataset
\State \Return trained policy $\pi$

\end{algorithmic}
\end{algorithm}

For RL training, we set two types of rewards. The first reward is for format, and the second reward is for answer correctness. For the format reward, we align with the action space and action format from WebVoyager. Each reward is 1.0, and if both are correct, the total reward is 2.0.

\begin{equation}
r_{\text{format}} =
\begin{cases}
1.0, & \text{if the action follows formats} \\
0.0, & \text{otherwise},
\end{cases}
\end{equation}

\begin{equation}
r_{\text{answer}} =
\begin{cases}
1.0, & \text{if the predicted answer is correct} \\
0.0, & \text{otherwise},
\end{cases}
\end{equation}

\begin{equation}
R_{rl} \;=\; r_{\text{format}} + r_{\text{answer}}.
\end{equation}

\section{PEEU Experiment Details}
\label{appendix:PEEU Dataset Details}
This section presents the implementation details of our experiments, including the data processing pipeline on the WebVoyager benchmark and the specific settings for In-Domain (ID) and Out-Of-Domain (OOD) evaluations for PEEU.

(1) We evaluate the planning capabilities of our models using the WebVoyager benchmark~\citep{he2024webvoyager}, which comprises real-world multimodal tasks across diverse categories such as shopping, research, coding, and travel. To ensure a stable evaluation environment, we exclude websites with strict access frequency limits (e.g., Cambridge Dictionary, Google Search, and Hugging Face). Consequently, our study focuses on the remaining accessible websites, which fully comply with terms of service~\citep{he2024webvoyager}.

(2) To rigorously assess cross-site generalization, we structure our dataset into distinct In-Domain (ID) and Out-Of-Domain (OOD) partitions. Allrecipes is utilized as the source for ID exploration and training. Specifically, we curate two datasets for the algorithm: ID Set (~0.1k tasks): Consists of approximately 100 tasks derived exclusively from Allrecipes. Supplementary OOD Set (~2k tasks): Consists of approximately 2,000 tasks collected from additional websites. Crucially, the websites used for the Supplementary OOD Set are distinct from the 7 held-out websites reserved strictly for testing. This setup allows us to train/explore on one specific site (and optionally augment with the 2k OOD pool) while testing on 7 completely unseen websites to evaluate zero-shot generalization. We filter out data with incorrect formats prior to usage. Following WebVoyager~\citep{he2024webvoyager} standard setting, for the experimental hyperparameters, the maximum exploration depth is set to 15 steps. The retrieval module employs all-roberta-large-v1~\citep{reimers-2020-multilingual-sentence-bert} for semantic matching.

\section{Definition Details}
\label{appendix:Definition Details}

In this section, we introduce and formalize the definitions of task planning, and then present the three levels of task planning granularity in this work, including low-level tasks, mid-level tasks, and high-level tasks. As well as the definitions of in-domain, out-of-domain and experience.

\paragraph{Task Planning Definition.}
The task planning is formally defined as a tuple~\citep{li2025perception,cao2025large,wei2025plangenllms}:
\begin{equation}
\mathcal{P} = \langle \mathcal{S}, \mathcal{A}, T, s_0, \mathcal{G} \rangle.
\end{equation}
Here, $\mathcal{S}$ is a set of environment states, $\mathcal{A}$ is a set of actions, $T: \mathcal{S} \times \mathcal{A} \rightarrow \mathcal{S}$ is a state transition function, $s_0 \in \mathcal{S}$ is an initial state, $\mathcal{G} \subseteq \mathcal{S}$ is a set of goal states. The objective is to find a sequence of actions $\langle a_0, a_1, \ldots, a_n \rangle$ that transforms the system from the initial state $s_0$ to a goal state $s_g \in \mathcal{G}$.

In the ReAct paradigm~\citep{yao2023react}, the objective is to output the next action given the task description, history, and current observation. This can be formally represented as:
\begin{equation}
a_t = \pi(d, \mathcal{H}_{0:t}, s_t).
\end{equation}
Here, $d$ is the task description, and $\mathcal{H}_{0:t} = \{(s_0, a_0), (s_1, a_1), \dots, (s_{t-1}, a_{t-1})\}$ is the history of state-action pairs up to time $t-1$, $s_t$ is the current observation, and $\pi$ is the planning policy that outputs the action $a_t$. Upon task completion, we obtain a trajectory $\tau = \{(s_0, a_0), (s_1, a_1), \dots, (s_n, a_n)\}$.

\paragraph{Low-level Task Definition.}
The low-level task is defined as a single-step task. It is also called the atomic-level task. For step $t$, the policy $\pi$ uses only the current low-level task description $d_{low}$ and the current observation $s_t$ to determine the next action:
\begin{equation}
a_t = \pi(d_{low}, s_t).
\end{equation}

\paragraph{Mid-level Task Definition.}
The mid-level task is defined as a multi-step subtask. For a subtask spanning steps $p$ to $q$, the policy $\pi$ uses the middle-level task description $d_{mid}$, the history $\mathcal{H}_{p:t}$ and the current observation $s_t$ to determine the next action:
\begin{equation}
a_t = \pi(d_{mid}, \mathcal{H}_{p:t}, s_t).
\end{equation}

\paragraph{High-level Task Definition.} 
The high-level task is defined as a long horizon, composed of a sequence of subtasks. For a long horizon task $0$ to $n$, the policy $\pi$ uses the high-level task description $d_{high}$, the history $\mathcal{H}_{0:t}$ and the current observation $s_t$ to determine the next action:
\begin{equation}
a_t = \pi(d_{high}, \mathcal{H}_{0:t}, s_t).
\end{equation}

\paragraph{In-Domain and Out-of-Domain.} 
For the TDHAF, ID evaluation uses test data from the same trajectories seen during post-training. The task description has been paraphrased, while OOD evaluation uses test data from entirely new websites not encountered during post-training.
For the PEEU, ID evaluation uses test data from the same websites seen during post-training, while OOD evaluation uses test data from entirely new websites not encountered during post-training.

\paragraph{Experience Definition.} 
As defined in ~\citet{silver2025welcome}, experience is defined as data produced through an agent's interactions with the environment. Subsequent work~\citep{cai2025building} further categorizes experiences into trajectories, knowledge and skills summarized from these trajectories. 
In this paper, we mainly refer to what is summarized from the trajectory as experience.

\section{TDHAF Prompt}
\label{appendix:TDHAF Prompt}

\begin{dialogbox}
Build Low Level Prompt for TDHAF

Your task is to generate task descriptions for CLICK/TYPE/SELECT an on-screen element.

Two screenshots are provided:

Current UI - Shows a interactive element (labeled "1") with a bounding box.

Post-interaction UI - Highlights changes after interaction (excluding bounding box disappearance).

Task: 
Purpose Clarity - Clearly define the purpose of the interaction with the UI element in both descriptions, ensuring they are functionally identical but phrased differently.

Ensure the two descriptions serve distinct contexts with no overlapping phrasing.

Action Consistency - Use only CLICK, TYPE, or SELECT as action types, with identical parameters in both descriptions (e.g., target element, input text, or selection option).

UI Change Focus - Describe only observable UI changes (e.g., new elements appearing, data updates, transitions) resulting from the action-avoid vague or future-oriented statements.

Training vs. Testing Wording - Paraphrase the purpose distinctly for training (instructional) and testing (validation) contexts while keeping functional outcomes identical.

Now, generate the two mission-style descriptions adhering to these rules. Only output the lists, nothing else.

The raw task is <task>.
\end{dialogbox}

\begin{dialogbox}
Build High Level Prompt for TDHAF

Please make this task more complex, but do not change the parameters in this task. Add more subtasks after this task, and rephrase the original task with synonymous expressions. This task and subsequent tasks can be combined into a more complex task. More complex means that the current task is a subtask in the middle, and then more subtasks are added before and after to merge into a more complex task. But don't describe the specific tasks in detail. Please output two task descriptions that are paraphrases of each other, in the form of a list of json. The key of the element is the string task, and the value is the task description.
The raw task is <task>.
\end{dialogbox}

\begin{dialogbox}
Inference Prompt for Multimodal-Mind2web for Agent

User:

<image>You are a web agent. 

Your task is: <task>

The history is: <history>. 

If you want to complete the task, you should output action CLICK/TYPE/SELECT, id and value in <answer> </answer> tags. 

Output the one bbox you should interact with in JSON format. 

Examples:

1. For clicking: <answer>{"action": "CLICK", "value": "" ,"id": 3}</answer>

2. For typing text: <answer>{"action": "TYPE","value": "example@email.com", "id": 5}</answer>

3. For selecting an option: <answer>{"action": "SELECT", "value": "United States","id": 2}</answer>
\end{dialogbox}

\section{TDHAF Data Construction}
\label{appendix:TDHAF Dataset Details}
\begin{figure}[h]
\begin{center}
\includegraphics[width=\linewidth]{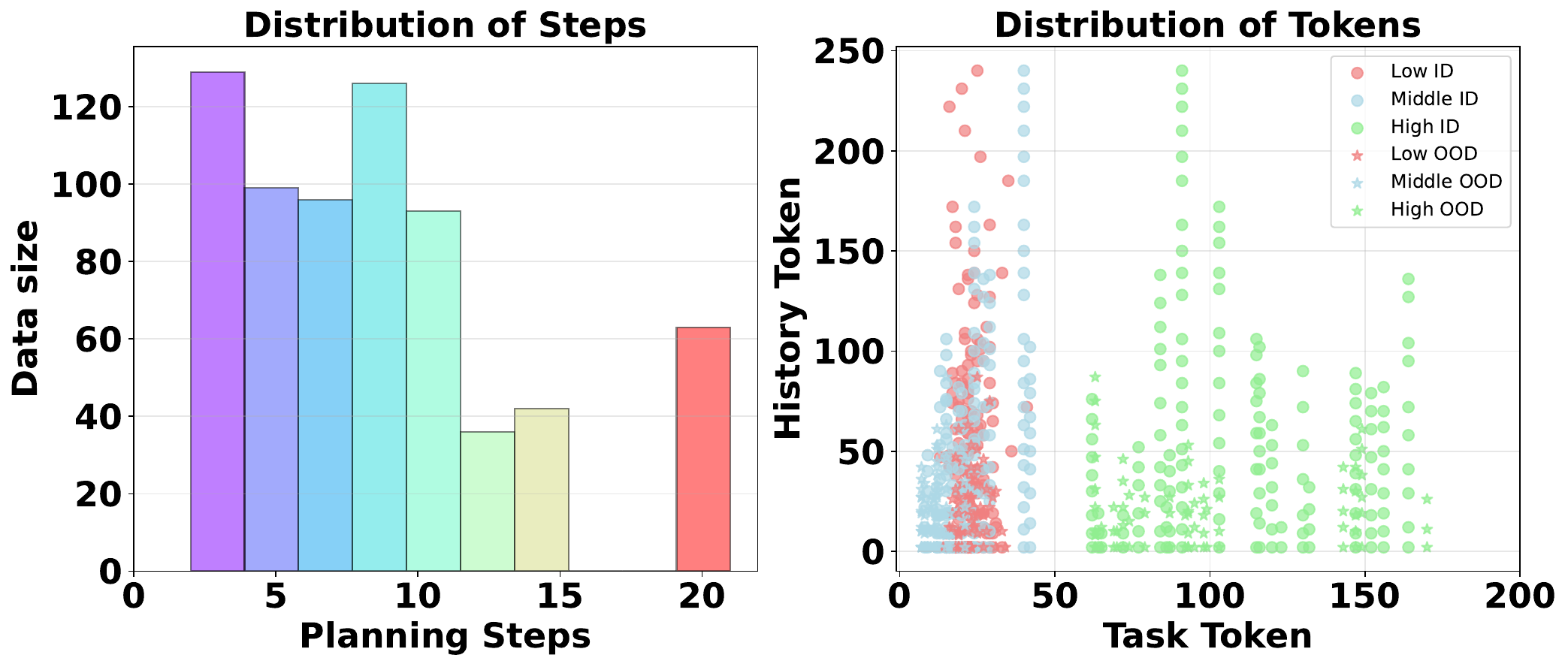}
\end{center}
\caption{Data Distribution for TDHAF.}
\label{fig:data_distribution}
\end{figure}

Raw data is collected from Multimodal-Mind2Web~\citep{deng2023mindweb,zheng2024seeact}. It is an offline human-expert-annotated gold trajectory dataset. Employing such a dataset for analysis offers more significant advantages, as it enables fine-grained examination of the model's behavior at the single-step level, including the target numbers, action types, action parameters. 
The in-domain test and train data come from the same trajectory, while the out-of-domain test data come from different trajectories of completely different websites. The in-domain training and test data are derived from the same trajectories, but the questions are rewritten. 
The training set has 616 samples, and the test set has 684 samples. The data statistics are shown in Figure~\ref{fig:data_distribution}. The data split is shown in Table~\ref{TDHAF:division}. The prompts for generating data are shown in Appendix~\ref{appendix:TDHAF Prompt}, which are the prompts for generating low-level tasks and high-level tasks by GPT-4o.

\begin{table}[h]
  \centering
  \caption{This table shows the TDHAF division of training and test sets for three generalization dimensions. ID indicates that training and test are derived from the same trajectory in the same websites, but the tasks are rewritten. OOD indicates they come from different trajectories across different websites. L denotes low-level tasks, M denotes mid-level tasks, H denotes high-level tasks.}
  \label{tab:generalization_dimensions}
  \scalebox{0.8}{
  \begin{tabular}{c c}
    \toprule
    \textbf{Training Set} & \textbf{Test Set} \\
    \midrule
    \multicolumn{2}{c}{\textbf{ID Bottom-up Generalization}} \\
    \midrule
    Train-ID-L & Test-ID-L, Test-ID-M, Test-ID-H \\
    Train-ID-M & Test-ID-M, Test-ID-H \\
    Train-ID-H & Test-ID-H \\
    \midrule
    \multicolumn{2}{c}{\textbf{ID Top-down Generalization}} \\
    \midrule
    Train-ID-L & Test-ID-L \\
    Train-ID-M & Test-ID-L, Test-ID-M \\
    Train-ID-H & Test-ID-L, Test-ID-M, Test-ID-H \\
    \midrule
    \multicolumn{2}{c}{\textbf{OOD Multi-level Generalization}} \\
    \midrule
    Train-ID-L & Test-OOD-L, Test-OOD-M, Test-OOD-H \\
    Train-ID-M & Test-OOD-L, Test-OOD-M, Test-OOD-H \\
    Train-ID-H & Test-OOD-L, Test-OOD-M, Test-OOD-H \\
    \bottomrule
  \end{tabular}
  }
\label{TDHAF:division}
\end{table}

\section{Generalization Distribution and Definition}
\label{appendix:Generalization Distribution and Definition}
Let the set of levels be
\begin{equation}
L = \{\text{low}, \text{middle}, \text{high}\}.
\end{equation}

For a sample $x$ at level $\ell \in L$, define an indicator
\begin{equation}
I(\ell, x) =
\begin{cases}
1, & \text{if the prediction at level $\ell$ is correct}, \\
0, & \text{otherwise}.
\end{cases}
\end{equation}

\paragraph{Good Generalization.}
The model is considered to generalize well at some level (low, middle, or high) if it predicts correctly not only at this level but also at the other two levels. That means correct at all three levels. Good generalization means successful generalization to other levels, the larger the better.
\begin{equation}
\begin{split}
\text{Good}(\ell, x) = 1 \\
\quad \text{if and only if} \quad I(\ell', x) = 1 \; \; \forall \ell' \in L.
\end{split}
\end{equation}

\paragraph{Bad Generalization.} 
The model is considered to generalize bad at some level if it is correct at this level, but at least one of the other two levels is wrong. Bad generalization means failure to fully generalize to other levels, the smaller the better.
\begin{equation}
\begin{split}
\text{Bad}(\ell, x) = 1
\quad \text{if and only if} \quad I(\ell, x) = 1 \; \\
\text{and} \; \exists \ell' \in L, \; \ell'\neq \ell \;\; \text{with } I(\ell', x) = 0.
\end{split}
\end{equation}

\paragraph{Coverage Percentage.} 
Among all samples that are predicted correctly at their own level, and these samples that are also correct at all three levels (i.e., that achieve good generalization) is called the \emph{coverage percentage}. 

Formally, let
\begin{equation}
G_\ell = \{ x \in S_\ell \mid \text{Good}(\ell, x) = 1 \}
\end{equation}
be the set of samples that are correctly predicted at level $\ell$ and also satisfy the good generalization condition. Here, $S_\ell$ denotes the set of all samples that are predicted correctly at level $\ell$, and $T$ denotes the entire test set. 

The coverage percentage at level $\ell$ is then defined as
\begin{equation}
\text{Coverage}(\ell) = \frac{|G_\ell|}{|T|} \times 100\% .
\end{equation}

\begin{figure*}[h]
\begin{center}
\includegraphics[width=\linewidth]{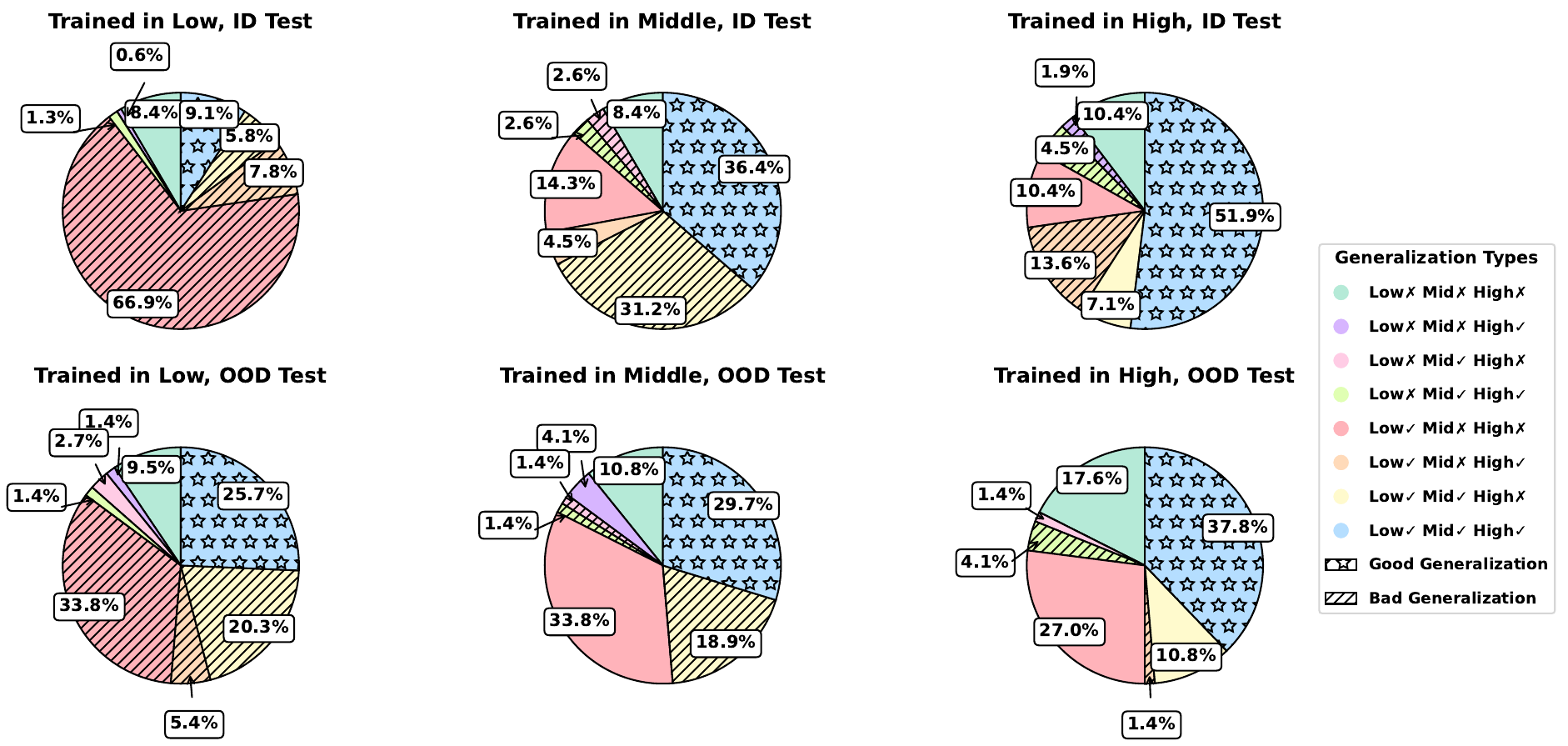}
\end{center}
\caption{Generalization Distribution Pie Chart for Qwen2.5-VL-7B. Good generalization means successful generalization to other levels, and the larger it is, the better. Bad generalization means failure to fully generalize to other levels, and the smaller it is, the better. The good generalization area expands from left to right, demonstrating high-level task training yields better generalization. Results for Qwen2.5-VL-7B.}
\label{fig:analysis_pie_7B}
\end{figure*}

\section{Usage of Chatgpt}

The use of ChatGPT is only limited to grammar checking and linguistic refinement.

\section{Training Reward Details}

\begin{figure*}
\begin{center}
\includegraphics[width=\linewidth]{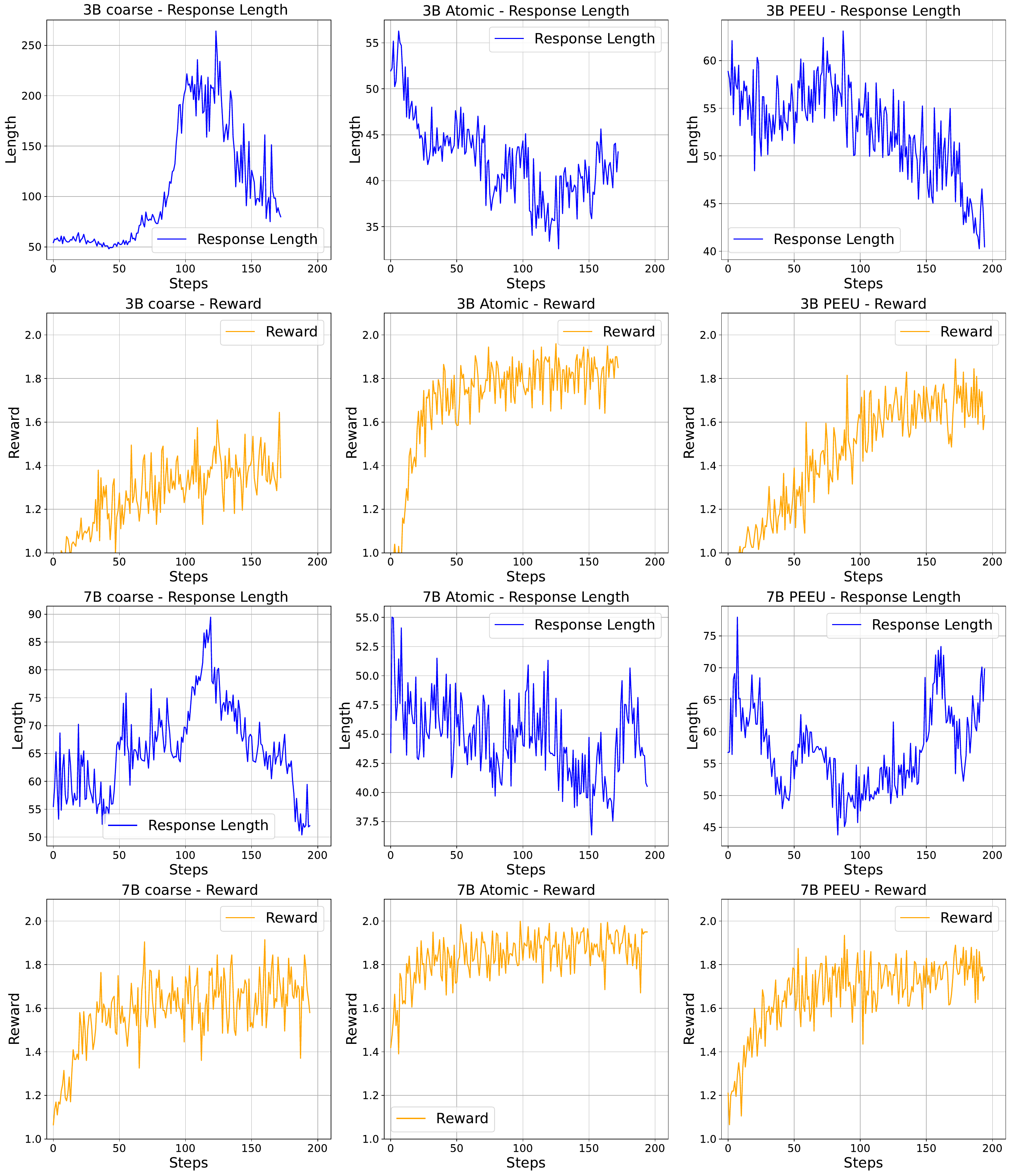}
\end{center}
\caption{RL Training Reward.}
\label{fig:rl_training_reward}
\end{figure*}

\end{document}